# FLaG: Frequency-Domain Latent-attention Gated Pooling for Token Aggregation

Kewei Li[1,2], Rongying Zhang[3], Xueli Wang[1,2], Xiwen Gong[4], Zhongjian Wang[5], Qiuchen Zhao[6,7], Lan Huang[1,2], Ruochi Zhang[1,2,#], Fengfeng Zhou[1,2,#].

1 College of Computer Science and Technology, Jilin University, Changchun, 130012, China.

2 Key Laboratory of Symbolic Computation and Knowledge Engineering of Ministry of Education, Jilin University, Changchun, 130012, China.

3 Institute for Quantitative and Computational Biology, University of California, Los Angeles, 90024, United States of America.

4 Greenwich High School, Greenwich, CT, 06830, United States of America.

5 BCPM Data Limited, Chengdu 610041, China.

6 Cancer Research UK Cambridge Centre and Department of Oncology, University of Cambridge, Cambridge, CB2 0XZ, UK

7 Department of Pathology, University of Cambridge, Cambridge, CB2 1QP, UK

# Correspondence may be addressed to Drs. Ruochi Zhang (zrc720@gmail.com), and Fengfeng Zhou (FengfengZhou@gmail.com or ffzhou@jlu.edu.cn).

## Abstract

Token aggregation converts token-level representations into fixed-dimensional sample representations, but most pooling methods operate only in the original token space. We introduce Frequency-Domain Latent-attention Gated Pooling (FLaG), a plug-in aggregation module that re-expresses encoder outputs in the Fourier domain before final pooling. FLaG represents the nonredundant rFFT spectrum through concatenated real and imaginary components, summarizes spectral tokens with learnable latent queries, derives a sample-conditioned channel gate, and reconstructs modulated token representations for downstream aggregation. We evaluate the same architecture across ESM2-based antimicrobial peptide (AMP) activity prediction, ResNet18 image classification on CIFAR-10 and CIFAR-100, and three RoBERTa-based language tasks. FLaG achieves the best macro-averaged Spearman correlation coefficient, RMSE, and Recall@50 across four AMP backbone-species settings and the highest top-1 accuracy on CIFAR 10. It also achieves the best mean results on five of seven language metrics, although mean pooling remains strongest on STSBenchmark. AMP-side mechanistic analyses reveal low-frequency prediction sensitivity across most encoder layers, with increased relative high-frequency sensitivity in the final layer, and pronounced peptide-specific positional responses. The residual gate broadly amplifies spectral channels while preserving the low-frequency-dominated energy profile, whereas latent cross-attention exhibits sample- and species-specific spectral allocation. Physicochemical probing further shows that the normalized DC representation preserves substantial global peptide information, while its association with antimicrobial activity differs between bacterial species. Overall, FLaG provides a transferable frequency-domain aggregation bias across protein, visual, and textual representations, with benefits that depend on the backbone and downstream task. Supplementary materials, source code, and data are available at https://www.healthinformaticslab.org/supp/ and https://github.com/Kewei2023/AMPCliff/tree/FLaG.

## 1 Introduction

Modern neural architectures across protein modeling [1-3], computer vision [4], and natural language processing [5] often face a common aggregation problem. An encoder produces a collection of token-level or spatial representations, while the downstream prediction head usually requires a fixed-dimensional sample representation [2, 6-8]. The final pooling operation therefore determines which information remains available after token reduction [2, 9-11]. In ESM2-based antimicrobial peptide models, residue embeddings must be aggregated into peptide-level representations for activity prediction [12, 13]. In vision models such as ResNet, spatial features are pooled before image classification [14]. In language models such as RoBERTa, token representations are reduced to sentence- or sequence-level vectors for downstream tasks [15].

Most pooling methods perform this reduction directly in the original token coordinates. Mean pooling, max pooling, and token selection apply fixed reduction rules, while attention-based methods learn data-dependent aggregation weights [7, 9, 16-18]. Latent-attention further compresses variable-size token sets through a small set of learnable queries [7, 16, 18, 19]. Pooling design can materially affect downstream performance across different representation architectures [20]. However, these approaches generally operate on the encoder representation in its original basis. They do not first re-express token-axis variation in an alternative coordinate system before information is compressed into a sample-level vector.

Frequency-domain representations provide such an alternative. Fourier coordinates separate slowly varying and rapidly varying token-axis patterns into explicit spectral components and provide a global description of variation across positions. Spectral analyses and frequency-domain architectures have been explored in language modeling [21], neural spectral analysis [22], time-series forecasting [23], multiscale representation learning [24], and visual recognition [25]. These studies demonstrate the utility of spectral representations, but they primarily use frequency-domain operations for token mixing, filtering, representation analysis, or backbone modeling. Their objective is not to develop a general pooling interface between an arbitrary encoder and its downstream prediction head. This distinction motivates frequency-domain re-expression as an intermediate stage for token aggregation.

Recent protein-specific pooling studies provide a complementary motivation for exploring alternative aggregation spaces. For example, EvoPool [26] augments residue-level protein language model embeddings with a fixed-size evolutionary anchor derived from homologous sequences and uses this external context to guide protein-level aggregation. Such strategies can be valuable when evolutionary information is readily available, but they also depend on protein-specific auxiliary information that may be sparse, unavailable, or undesirable at inference time. This limitation is particularly relevant to antimicrobial peptide prediction, where short sequences may provide limited homologous context. It also prevents the same aggregation mechanism from being transferred directly to non-protein representations. These considerations motivate an internal alternative that operates only on encoder outputs and does not require homologs, multiple sequence alignments, structural annotations, or other domain-specific anchors.

We therefore introduce Frequency-Domain Latent-attention Gated Pooling (FLaG), a plug-in module that re-expresses encoder outputs in spectral coordinates before final token reduction. FLaG applies rFFT along the token axis and represents the nonredundant complex spectrum through concatenated real and imaginary components. Learnable latent queries summarize the resulting frequency tokens [16, 18], and a sample-conditioned channel gate modulates the spectral representation [27]. The inverse transform then reconstructs modulated token representations for standard pooling. FLaG operates after the encoder and does not require modification of the backbone or domain-specific external anchors. We evaluate the same core architecture on ESM2-based antimicrobial peptide activity prediction [12, 28], ResNet18 image classification on CIFAR-10 and CIFAR-100 [14, 29], and RoBERTa-based language tasks drawn from MTEB [15, 30]. Beyond cross-domain benchmarking, we use AMP activity prediction as a mechanistic case study because residue-level representations and sequence-derived physicochemical properties allow the aggregation process to be examined at positional, spectral, and property levels.

This study makes three main contributions:

- We introduce FLaG, a plug-in pooling framework that combines spectral re-expression, latent-query summarization, sample-conditioned channel gating, inverse reconstruction, and final token aggregation.
- We evaluate the same core FLaG architecture across protein regression, image classification, and language tasks. The results characterize where frequency-domain aggregation is beneficial and show that its effect depends on the backbone and downstream objective.
- We use AMP activity prediction as a mechanistic case study to examine how FLaG processes encoded token representations. Five complementary analyses characterize positional sensitivity, layer-wise frequency dependence, spectral modulation, latent-query allocation, and physicochemical information in low-frequency components. The fifth analysis comprises two linked parts: physicochemical-property encoding and spectral sensitivity, followed by species-specific property–activity associations.

# 2 Related Work

## 2.1 Token Pooling Methods

Mean pooling treats all tokens as equally informative. This uniform weighting can dilute rare but functionally critical residues or sequence elements [31]. It may also overlook higher-order structure in the protein embedding space [26]. Max pooling takes the strongest activation along each feature dimension. It can preserve salient local responses, but it may also become sensitive to outlier activations that are not task-relevant [31]. Last-token selection assumes that sequence-level information is concentrated in the terminal hidden state [32]. This assumption is reasonable for autoregressive decoders, where the final valid token summarizes the preceding context [32]. It is less suitable for bidirectional encoders, which usually rely on a special summary token or explicit token pooling [5]. Last-token selection may also introduce recency bias when the final token dominates the representation. In contrast, simple averaging may dilute phrase-level or motif-level signals [16].

Attention pooling improves this fixed reduction by learning token-level importance weights in the original token domain [9, 33]. However, this flexibility comes with additional parameters and may be sensitive to irrelevant instances when supervision is weak, or dataset-specific token responses [34, 35]. Multi-Layer Trainable Pooling further broadens the aggregation source by combining hidden-layer representations before final token pooling [17]. This design can exploit complementary information encoded at different depths, since shallow, intermediate, and deep layers may capture different levels of sequence structure and task-relevant abstraction [36]. However, it introduces an additional trainable layer-fusion module, and its benefit appears to be task dependent.

The method most closely related to FLaG is latent-attention [7, 16, 18, 19]. It extends this idea through multiple learnable query vectors, which allow a compact set of latent summaries to attend to the input token set [7]. This design can capture complementary evidence patterns and provides a flexible bottleneck for variable-length sequences [7, 18, 19]. A key limitation is its latent bottleneck. When the number of input tokens or the amount of information to be compressed increases, a correspondingly larger latent set may be required [37]. Latent-attention also remains a token-domain aggregation strategy. It can learn which tokens to emphasize, but it does not explicitly decompose global sequence trends, local motif-level variations, or frequency-domain perturbations.

## 2.2 Frequency-Domain Methods in Deep Learning

Frequency-domain modeling offers a different view of representation structure [21]. Spectral analyses of neural networks have shown that deep models often fit low-frequency components before high-frequency components [22]. Graph neural networks provide another example, where graph convolution can be derived from spectral filters on the graph Laplacian [38]. In time-series modeling, frequency-enhanced Transformers use Fourier-domain representations to capture global and periodic signals [23]. In visual recognition, GFNet performs global filtering in the frequency domain [25]. These studies support the value of spectral representations, but they address different modeling levels. Graph spectral methods define convolution on graph structure. Frequency-enhanced forecasting modifies the sequence model itself. FLaG in this study instead applies spectral re-expression after the encoder and before final pooling. It combines FFT-based token re-expression, latent-query spectral summarization, and channel-wise spectral gating as a plug-in aggregation module.

## 2.3 Protein Representation Learning

Protein language models generate residue-level embeddings that encode information about protein sequence, structure, and function [12]. For peptide and protein prediction, these residue embeddings must be reduced to a sequence-level representation. The choice of pooling can affect downstream activity prediction, especially when small sequence changes alter function [26, 31, 39]. Recent protein-specific pooling work addresses this issue by adding biological context to the aggregation step. EvoPool constructs a fixed-size evolutionary anchor from homologous sequences and uses this anchor to guide protein-level aggregation [26]. Other protein pooling methods use convolutional operators or internal transformer attention to derive residue weights for sequence-level prediction [31, 39]. These methods are useful when homologs, graph-derived signals, or residue-level biological priors are available. They are less suitable for images and text. They can also be difficult to use in antimicrobial peptide prediction when homologous context is sparse or unavailable at inference time. FLaG targets this gap through an internal frequency-domain representation. It does not require homologs, multiple sequence alignments, structural annotations, or external anchors.

# 3 Methods

## 3.1 Problem Formulation

Let $\mathbf{X} \in \mathbb{R}^{T\times D}$ denote a sequence of token representations produced by an upstream encoder, where $T$ is the sequence length and $D$ is the hidden dimension. A pooling module maps $\mathbf{X}$ to a fixed-length vector $\mathbf{z} \in \mathbb{R}^{D}$ for sample-level prediction. When variable-length sequences are padded to a common length, an optional binary mask $m \in \{0,1\}^{T}$ identifies valid token positions. The pooling operation should prevent padded positions from contributing to the final representation. It should also compress the token sequence while preserving information that is relevant to the downstream task.

Conventional pooling methods perform this reduction directly in the token domain. FLaG instead introduces an intermediate frequency-domain representation before final pooling. This design allows token variation across positions to be summarized in spectral space, modulated through a learned gate, and then mapped back to the token domain for prediction.

## 3.2 Overview

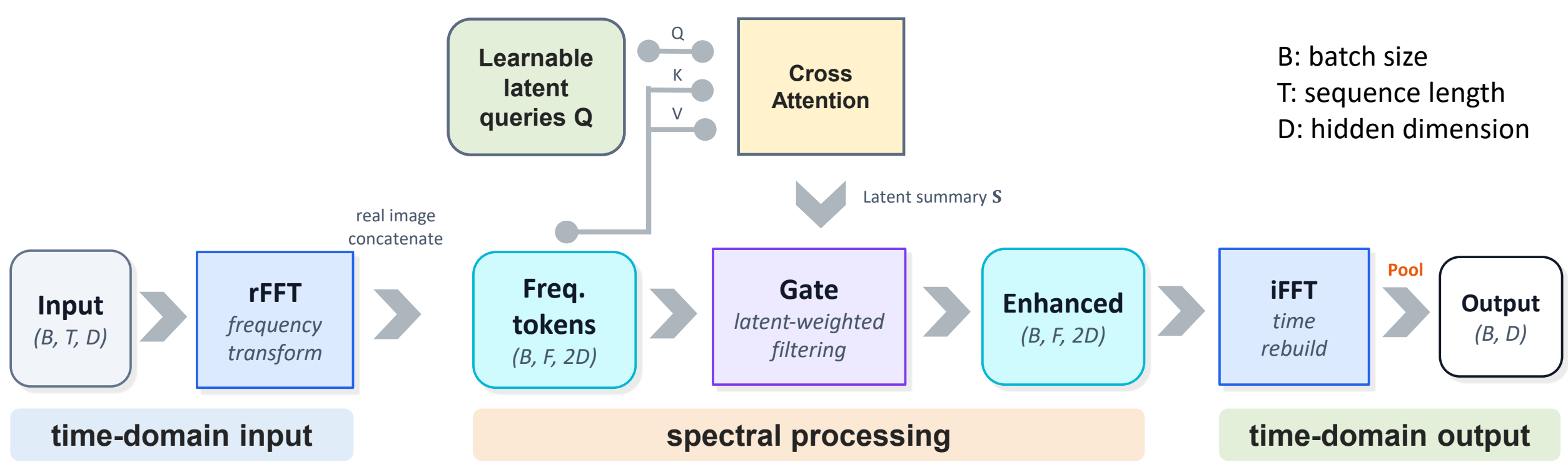

**Figure 1. Architecture of Frequency-Domain Latent-attention Gated Pooling.** Token representations are transformed with real-to-complex Fast Fourier Transform (rFFT) along the token dimension. The resulting frequency tokens are formed by concatenating the real and imaginary components of the Fourier coefficients. Learnable latent queries summarize spectral information and produce a gate context. A channel-wise spectral gate is shared across frequency bins and rescales the frequency representation before inverse reconstruction and final pooling.

FLaG contains four stages: frequency transformation, latent-attention, frequency gating, and time-domain pooling. Figure 1 shows the complete pipeline.

## 3.3 FLaG Processing Pipeline

Existing pooling methods usually aggregate encoder outputs directly in the token domain [9, 12, 15], which may compress token-level variation before alternative structural views are considered. FLaG instead re-expresses the encoder output in spectral coordinates, summarizes the resulting frequency tokens through learnable latent queries, and applies sample-conditioned spectral modulation before final pooling. This formulation does not assume that any predefined frequency band is universally informative. Rather, downstream supervision determines how the spectral representation contributes to aggregation.

### 3.3.1 Frequency Transformation

FLaG first applies the mask to the encoder output and then transforms the masked sequence along the token dimension. Let $\mathbf{X} \in \mathbb{R}^{T \times D}$ denote the padded token representation in a mini-batch, where $T$ is the padded sequence length and $D$ is the hidden dimension. Let $\mathbf{m} \in \{0,1\}^{T}$ denote the optional validity mask. FLaG applies a real-to-complex fast Fourier transform (rFFT):

$$\hat{\mathbf{X}} = \text{rFFT}(\mathbf{X} \odot \mathbf{m}) \in \mathbb{C}^{F \times D}, F = \lfloor T/2 \rfloor + 1 \quad (1)$$

Here, $F$ denotes the number of nonredundant frequency bins. In variable-length batches, sequences are padded to a shared length before transformation. The mask prevents padded positions from contributing to the transformed signal. Since the masked encoder output is real-valued, its full Fourier spectrum satisfies conjugate symmetry. The negative-frequency coefficients are therefore redundant and can be omitted without losing the information required for inverse reconstruction. Each retained coefficient depends on all token positions, so the spectral representation provides a global coordinate system for sequence summarization before final reduction. The coefficient at k=0 is the zero-frequency component, and it is also known as the Direct Current (DC) component. It represents the sequence-level mean component up to the selected normalization convention.

The complex spectrum is then converted into a real-valued tensor by concatenating the real and imaginary components:

$$\boldsymbol{\mathcal{F}} = \left[\text{Re}(\hat{\mathbf{X}}) \parallel \text{Im}(\hat{\mathbf{X}})\right] \in \mathbb{R}^{F \times 2D} \quad (2)$$

Although the input sequence is real-valued, the retained rFFT coefficients are generally complex. The real and imaginary components encode the cosine- and sine-basis projections, respectively, with the imaginary component retaining phase-sensitive information. Their concatenation keeps both components of the Fourier coefficients and allows standard real-valued neural layers to operate on the spectral tokens.

### 3.3.2 Latent-attention in Frequency Space

Let $L$ denote the number of learnable latent queries, which is typically smaller than the number of frequency tokens, $L < F$. We define the query matrix as $\boldsymbol{Q} \in \mathbb{R}^{L\times 2D}$. For each sample, the same latent queries attend to the frequency-token matrix $\boldsymbol{\mathcal{F}} \in \mathbb{R}^{F\times 2D}$. The frequency tokens serve as both keys and values. This cross-attention maps the $\boldsymbol{\mathcal{F}}$ frequency tokens to a fixed set of $L$ latent representations [18, 19, 40]. This design follows the latent-bottleneck principle used in Perceiver and Perceiver IO.

$$\boldsymbol{\mathcal{A}} = \text{MHA}(\boldsymbol{Q}, \boldsymbol{\mathcal{F}}, \boldsymbol{\mathcal{F}}) \in \mathbb{R}^{L\times 2D}. \tag{3}$$

Here, MHA() denotes multi-head cross-attention, and $\boldsymbol{\mathcal{A}}$ denotes its output. Each row of $\boldsymbol{\mathcal{A}}$ contains the spectral evidence retrieved by one latent query. No predefined correspondence exists between a query and an individual frequency bin. The queries are therefore treated as learned spectral summary slots rather than explicit frequency detectors.

FLaG then applies two post-norm residual blocks [40-42]. The first block combines the cross-attention output with the original latent queries:

$$\boldsymbol{\mathcal{H}} = \text{LayerNorm}\big(\boldsymbol{Q} + \text{Dropout}(\boldsymbol{\mathcal{A}})\big) \in \mathbb{R}^{L\times 2D} \tag{4}$$

The second block refines each latent representation with a position-wise feed-forward network:

$$\boldsymbol{\mathcal{S}} = \text{LayerNorm}\Big(\boldsymbol{\mathcal{H}} + \text{Dropout}\big(\text{FFN}(\boldsymbol{\mathcal{H}})\big)\Big) \in \mathbb{R}^{L\times 2D} \tag{5}$$

Here, $\boldsymbol{\mathcal{H}}$ is the normalized attention representation, and $\boldsymbol{\mathcal{S}}$ is the refined latent tensor. The $\mathrm{FFN}$ acts independently on each latent row. It contains a $2D - \text{to} - 2D$ linear layer, a $\mathrm{GELU}$ activation, dropout, and a second $2D - \text{to} - 2D$ linear layer. A separate dropout operation is applied to the $\mathrm{FFN}$ output before residual addition. The latent width remains $2D$ throughout this stage. The next stage uses $\boldsymbol{\mathcal{S}}$ to construct the sample-conditioned channel gate. The post-norm equations and $\mathrm{FFN}$ dimensions match the released implementation.

The latent-query bottleneck therefore summarizes the full spectrum through a compact learned interface without requiring hand-defined frequency bands.

### 3.3.3 Latent-Conditioned Spectral Gating

The refined latent tensor $\mathbf{S} \in \mathbb{R}^{L\times 2D}$ contains $L$ spectral summaries for each sample. FLaG averages these summaries across the latent dimension:

$$\bar{\boldsymbol{\mathcal{S}}} = \frac{1}{L}\sum_{l=1}^{L} \boldsymbol{\mathcal{S}}_{l,:} \in \mathbb{R}^{2D}. \tag{6}$$

Here, $\boldsymbol{\mathcal{S}}_{l,:}$ denotes the $l$-th latent vector, and $\bar{\boldsymbol{\mathcal{S}}}$ provides a sample-level summary of the frequency-token representation. A two-layer multilayer perceptron maps this summary to a channel-wise gate:

$$\mathbf{g} = \sigma\big(\text{MLP}(\bar{\boldsymbol{\mathcal{S}}})\big) \in (0,1)^{2D}. \tag{7}$$

Here, $\sigma$ denotes the element-wise sigmoid function. The $\mathrm{MLP}$ contains two $2D - \text{to} - 2D$ linear layers. A $\mathrm{GELU}$ activation and dropout are placed between them. This construction follows the general channel-recalibration principle of squeeze-and-excitation networks [27], but FLaG derives the gate from latent spectral summaries rather than spatially pooled convolutional features.

The gate is shared across frequency bins and is applied to the concatenated real and imaginary spectral channels through residual multiplicative modulation:

$$\widetilde{\boldsymbol{\mathcal{F}}}_{f,:} = \boldsymbol{\mathcal{F}}_{f,:} \odot (1_{2D} + \mathbf{g}), f \in \{0, \ldots, F-1\}, \widetilde{\boldsymbol{\mathcal{F}}} \in \mathbb{R}^{F\times 2D}. \tag{8}$$

Here, $\odot$ denotes element-wise multiplication, and $\mathbf{1_{2D}}$ is an all-ones vector of length $2D$. The same gate $\mathbf{g}$ is applied to every frequency bin. However, its values depend on the sample-specific latent summary in Eq. 6. Accordingly, the gate performs latent-conditioned spectral-channel recalibration rather than explicit selection of individual frequency bins.

Because each gate value lies between zero and one, the effective scaling factor $1 + \mathbf{g}$ lies between one and two. The modulation therefore approaches the identity transformation when $\mathbf{g}$ is close to zero and amplifies the corresponding spectral channel when $\mathbf{g}$ is larger. It does not suppress a channel below its original scale. The gated representation $\widetilde{\boldsymbol{\mathcal{F}}}$ is subsequently converted back to a complex spectrum for inverse reconstruction.

### 3.3.4 Time-Domain Reconstruction and Final Pooling

The gated frequency representation $\widetilde{\boldsymbol{\mathcal{F}}} \in R^{F\times 2D}$ is first partitioned along its feature dimension into real and imaginary components:

$$\widetilde{\boldsymbol{\mathcal{F}}} = \widetilde{\boldsymbol{\mathcal{F}}}_R \parallel \widetilde{\boldsymbol{\mathcal{F}}}_I, \widetilde{\boldsymbol{\mathcal{F}}}_R \in R^{F\times D}, \widetilde{\boldsymbol{\mathcal{F}}}_I \in \mathbb{R}^{F\times D} \tag{9}$$

Here, $\widetilde{\boldsymbol{\mathcal{F}}}_R$ and $\widetilde{\boldsymbol{\mathcal{F}}}_I$ contain the gated real and imaginary Fourier coefficients, respectively. They are recombined into a one-sided complex spectrum. FLaG then applies the inverse real Fourier transform along the frequency axis:

$$\widetilde{\mathbf{X}} = irFFT\big(\widetilde{\boldsymbol{\mathcal{F}}}_R + i\widetilde{\boldsymbol{\mathcal{F}}}_I\big) \in \mathbb{R}^{T\times D} \tag{10}$$

The argument $n = T$ fixes the reconstructed sequence length to the padded input length. The inverse transformation returns a real-valued token representation $\widetilde{\mathbf{X}}$. The validity mask remains active during final pooling, so padded positions do not contribute to the sample-level representation:

$$\mathbf{z} = \mathrm{Pool}\left(\widetilde{\mathbf{X}}, \mathbf{m}\right) \in \mathbb{R}^D \tag{11}$$

The implementation supports masked max pooling and masked mean pooling. Max pooling is used as the default final aggregation in the main FLaG experiments. The domain-dependent post-pooling transformation is defined as

$$\mathcal{N}_\tau(\mathbf{z}) = \mathrm{LayerNorm}(\mathbf{z}) \textit{ if } \tau \in \{\textit{image}, \textit{text}\}, \mathcal{N}_\tau(\mathbf{z}) = \mathbf{z} \textit{ if } \tau = \textit{AMP} \tag{12}$$

Here, $\tau$ denotes the application domain. For image and text tasks, LayerNorm is applied after pooling to recenter and rescale the pooled representation [40, 41] before the final projection and prediction head [4, 42, 43]. For AMP regression, this post-pooling normalization is omitted because it reduced predictive performance in the corresponding empirical comparison.

The pooling-module output is

$$\mathbf{z}_{\mathrm{out}} = \mathbf{W}_{\mathrm{out}} \mathrm{Dropout}\left(N_\tau(\mathbf{z})\right) + \mathbf{b}_{\mathrm{out}} \tag{13}$$

Here, $\mathbf{W}_{\mathrm{out}} \in \mathbb{R}^{D \times D}$ is the learnable parameter of the output projection. The resulting vector is passed to the task-specific prediction head.

# 4 Experiments

## 4.1 Experimental Setup

### 4.1.1 Evaluation Tasks, Datasets, and Metrics

We evaluate FLaG across three domains: antimicrobial peptide activity prediction, image classification, and language representation tasks. This cross-domain design tests the same aggregation framework on protein, visual, and textual token representations.

Antimicrobial Peptide Activity Prediction. We use ESM2-8M and ESM2-35M [12] as the backbone models and evaluate quantitative antimicrobial activity prediction for *E. coli* and *S. aureus*. Dataset construction and activity-cliff-aware partitioning follow AMPCliff [28]. The AC split targets generalization under activity-cliff constraints rather than purely random partitioning. AMPCliff represents activity through -$\log_{10}$(MIC[M]), where a larger value indicates stronger antimicrobial activity. We report Spearman correlation, Recall@50, and RMSE. Recall@50 counts the overlap between the 50 peptides with the highest predicted activity and the 50 peptides with the highest reference activity, as defined in AMPCliff [28]. RMSE complements the rank-based and retrieval-based measures because it quantifies prediction error on the regression target scale.

Image Classification. We adopt an image-pooling protocol based on Universal Pooling [20]. ResNet18 [14] serves as the backbone. Its final spatial feature map is reshaped into a token sequence before aggregation, which provides a common input format for all pooling methods. We evaluate CIFAR-10 and CIFAR-100 [29]. We report top-1 and top-5 classification accuracy. For each reported metric, checkpoint selection uses the corresponding validation accuracy.

Language Representation Tasks. We use RoBERTa-base [15] and evaluate three English datasets included in MTEB [30]: IMDB for sentiment classification, STSBenchmark for semantic textual similarity, and SprintDuplicateQuestions for duplicate-question detection. Several pooling modules in our comparison contain trainable parameters. We therefore evaluate them under supervised downstream adaptation rather than as a standard MTEB leaderboard submission [16, 44-46]. This experimental design allows each trainable pooling module to receive matched downstream supervision. We report accuracy and F1 for IMDB, accuracy, average precision, and F1 for SprintDuplicateQuestions, and Spearman and Pearson correlation for STSBenchmark.

### 4.1.2 Pooling Baselines

We compare FLaG with six pooling baselines that span fixed, attention-based, distributional, and multi-layer aggregation strategies. These baselines are mean pooling [8], max pooling [47], last-token selection (last) [16], attention pooling (attn) [9], latent-attention (latent_attn) [7] and Multi-Layer Trainable Pooling (MLTP) [17]. All pooling methods receive the same task-specific backbone representations and follow the matched training protocols described below. For the language tasks, we additionally include first-token selection as a RoBERTa-specific fixed-token baseline.

### 4.1.3 Training Protocols and Model Selection

All main experiments are repeated over 10 random seeds. FLaG uses eight latent queries, four attention heads, a dropout rate of 0.1, the residual gate connection described in the Methods section, and max pooling for final time-domain aggregation. These core FLaG hyperparameters remain fixed across the three domains. No domain-specific tuning is performed for these settings. The pooling module is inserted immediately before the task-specific prediction head. Within each task, all pooling methods use the same data partition, backbone, optimization budget, and checkpoint-selection rule. The optimizer family and learning-rate schedule also remain matched within each task.

The AMP protocol evaluates ESM2-8M (t6) and ESM2-35M (t12) on the fixed *E. coli* and *S. aureus* AC splits. The maximum peptide sequence length is set to 28. Models optimize mean squared error, and checkpoint selection uses validation Spearman correlation.

The Image protocol evaluates CIFAR-10 and CIFAR-100 with a CIFAR-adapted ResNet18. We use 45,000 images for training and 5,000 for validation from the official training set. The official 10,000-image test set is retained for final evaluation. Models are trained for 250 epochs with the standard CIFAR augmentation used in our implementation and SGD optimization. Checkpoints are selected from validation performance.

The Language protocol uses IMDB and STSBenchmark for the end-to-end RoBERTa-base fine-tuning for three epochs. Without specific mention in Supplementary Material S1, the learning rate is 1e-3 for the pooling and prediction layers and 1e-5 for the RoBERTa backbone by default. IMDB uses a 90:10 partition of its official training set for training and validation. STSBenchmark uses its official development set for validation. SprintDuplicateQuestions does not provide an official training split in the adopted task configuration. We therefore freeze the RoBERTa backbone and adapt only the pooling module and the lightweight cosine-logit head. Adaptation uses a 90:10 partition of the official validation set, and final evaluation uses the official test set. Additional implementation details are provided in Supplementary Material S1.

## 4.2 Main Results

### 4.2.1 Antimicrobial Peptide Activity Prediction

**Table 1. Antimicrobial peptide activity prediction across ESM2 backbones and bacterial species.** Values are reported as mean ± standard deviation over 10 random seeds. Bold values indicate the best mean for each metric within each backbone-species setting. Higher values are better for Spearman correlation and Recall@50, whereas lower values are better for RMSE. For each metric, **Avg.** summarizes performance across all evaluated backbone–species settings. The **mean** subcolumn reports the arithmetic mean of the setting-specific mean values, whereas the **Std.** subcolumn reports the arithmetic mean of their corresponding standard deviations across random seeds.

| Pooling | ESM2-8M*E.coli* | | | ESM2-8M*S.aureus* | | | ESM2-35M*E.coli* | | |
|---|---|---|---|---|---|---|---|---|---|
| | Spearman↑ | RMSE↓ | Recall@50↑ | Spearman↑ | RMSE↓ | Recall@50↑ | Spearman↑ | RMSE↓ | Recall@50↑ |
| **mean** | 0.542±0.02 | 0.578±0.01 | 17.700±2.26 | 0.570±0.03 | 0.577±0.02 | 16.200±2.82 | 0.562±0.01 | 0.555±0.01 | 19.300±3.06 |
| **max** | 0.540±0.02 | 0.577±0.03 | 17.000±2.49 | 0.591±0.02 | 0.552±0.02 | **18.400±2.12** | **0.572±0.01** | 0.547±0.01 | 20.330±2.35 |
| **attn** | 0.525±0.02 | 0.599±0.01 | **21.200±1.32** | 0.374±0.03 | 0.656±0.01 | 13.100±2.28 | 0.536±0.02 | 0.588±0.02 | **21.600±2.01** |
| **last** | 0.551±0.01 | 0.566±0.01 | 17.000±2.83 | 0.576±0.02 | 0.563±0.01 | 18.200±1.64 | 0.561±0.02 | 0.554±0.01 | 20.800±1.30 |
| **MLTP** | 0.516±0.02 | 0.608±0.01 | 19.000±3.94 | 0.419±0.02 | 0.646±0.01 | 12.700±1.89 | 0.535±0.02 | 0.595±0.01 | 21.000 ± 2.21 |
| **latent_attn** | 0.541±0.02 | 0.577±0.02 | 16.700±3.71 | 0.570±0.02 | 0.570±0.01 | 15.500±2.17 | 0.561±0.03 | 0.561±0.02 | 19.290±2.63 |
| **FLaG** | **0.554±0.01** | **0.562±0.01** | 18.600±2.50 | **0.603±0.02** | **0.545±0.00** | 18.000±1.76 | 0.570±0.02 | **0.546±0.01** | 20.900±2.08 |

| Pooling | ESM2-35M*S.aureus* | | | Avg.Spearman↑ | | Avg.RMSE↓ | | Avg.Recall@50↑ | |
|---|---|---|---|---|---|---|---|---|---|
| | Spearman↑ | RMSE↓ | Recall@50↑ | mean | Std. | mean | Std. | mean | Std. |
| **mean** | 0.580±0.02 | 0.559±0.01 | 17.700±2.21 | 0.563 | 0.02 | 0.567 | 0.01 | 17.725 | 2.59 |
| **max** | 0.593±0.01 | 0.548±0.01 | 18.300±1.25 | 0.574 | 0.02 | 0.556 | 0.01 | 18.508 | 2.05 |
| **attn** | 0.412±0.02 | 0.619±0.01 | 16.100±1.52 | 0.462 | 0.02 | 0.615 | 0.01 | 18.000 | 1.78 |
| **last** | **0.596±0.02** | **0.544±0.01** | 17.600±0.55 | 0.571 | 0.02 | 0.557 | 0.01 | 18.400 | 1.58 |
| **MLTP** | 0.446±0.02 | 0.617±0.01 | 17.500±2.64 | 0.479 | 0.02 | 0.617 | 0.01 | 17.550 | 2.67 |
| **latent_attn** | 0.582±0.01 | 0.557±0.01 | 17.300±2.21 | 0.563 | 0.02 | 0.566 | 0.01 | 17.197 | 2.68 |
| **FLaG** | 0.582±0.02 | 0.554±0.01 | **18.600±1.35** | **0.577** | **0.02** | **0.552** | **0.01** | **19.025** | **1.92** |

Table 1 summarizes AMP activity prediction with ESM2-8M and ESM2-35M across *E. coli* and *S. aureus*. With ESM2-8M, FLaG provides the strongest overall profile across the two species. It achieves the highest Spearman correlation for both *E. coli* and *S. aureus*, with values of 0.554 and 0.603, respectively. FLaG also yields the lowest RMSE in both settings, at 0.562 and 0.545. Relative to mean pooling, these values correspond to RMSE reductions of 0.016 for *E. coli* and 0.032 for S. aureus. For E. coli, attn achieves the highest Recall@50 of 21.2, while FLaG reaches 18.6. For S. au*reus*, max pooling gives the highest Recall@50 at 18.4, while FLaG remains close at 18.0. Thus, the ESM2-8M results show that FLaG improves both rank correlation and regression error across the two species while maintaining competitive top-candidate recovery.

With ESM2-35M, the advantage of FLaG becomes more metric-dependent. For E. coli, FLaG obtains the lowest RMSE of 0.546 and a Recall@50 of 20.9, while attn achieves the highest Recall@50 of 21.6. Max pooling gives a slightly higher Spearman correlation, at 0.572 compared with 0.570 for FLaG. For *S. aureus*, last-token pooling achieves the highest Spearman correlation of 0.596 and the lowest RMSE of 0.544. FLaG reaches 0.582 and 0.554 on these two metrics, respectively, but achieves the highest Recall@50 at 18.6. These results show that the FLaG advantage is less uniform with ESM2-35M than with ESM2-8M. The preferred pooling strategy depends more strongly on the metric and bacterial species in this setting.

The macro-average across all four backbone-species settings provides an overall comparison. FLaG achieves the highest average Spearman correlation at 0.577, the lowest average RMSE at 0.552, and the highest average Recall@50 at 19.025. Compared with max pooling which provides the strongest aggregate baseline, FLaG increases Spearman correlation by 0.003, reduces RMSE by 0.004, and increases Recall@50 by 0.518. The differences relative to mean pooling are larger, with a 0.014 increase in Spearman correlation, a 0.015 reduction in RMSE, and a 1.300 increase in Recall@50. FLaG also outperforms the latent-attention baseline on all three aggregate metrics. Overall, FLaG does not dominate every individual backbone-species-metric combination, but it provides the strongest aggregate performance across rank correlation, regression error, and top-candidate recovery.

### 4.2.2 Image Classification

**Table 2. Image classification performance of pooling methods with ResNet18 on CIFAR-10 and CIFAR-100.** Results are reported as mean ± standard deviation over 10 random seeds. Bold values indicate the best mean performance in each column. Each test metric is obtained from the checkpoint selected according to the corresponding validation metric. Avg. Acc denotes the average of the four mean accuracy values, and Avg. Std denotes the average of their corresponding standard deviations.

| Pooling | CIFAR-10 (top-1) | CIFAR-10 (top-5) | CIFAR-100 (top-1) | CIFAR-100 (top-5) | Avg. Acc | Avg. Std |
|---|---|---|---|---|---|---|
| **mean** | 0.9594±0.0010 | 0.9986±0.0006 | 0.7678±0.0031 | 0.9325±0.0014 | 0.9146 | 0.0015 |
| **max** | 0.9577±0.0019 | 0.9986±0.0004 | 0.7661±0.0036 | **0.9342±0.0025** | 0.9142 | 0.0021 |
| **attn** | 0.9460±0.0006 | 0.9981±0.0003 | 0.7305±0.0023 | 0.9132±0.0018 | 0.89695 | 0.0012 |
| **last** | 0.9589±0.0012 | 0.9986±0.0005 | 0.7514±0.0043 | 0.9259±0.0017 | 0.9087 | 0.0019 |
| **MLTP** | 0.9581±0.0014 | **0.9988±0.0004** | **0.7759±0.0032** | 0.9250±0.0019 | 0.91445 | 0.0017 |
| **latent_attn** | 0.9596±0.0012 | 0.9987±0.0003 | 0.7571±0.0036 | 0.9219±0.0030 | 0.9093 | 0.0020 |
| **FLaG** | **0.9601±0.0017** | 0.9986±0.0003 | 0.7720±0.0014 | 0.9313±0.0012 | **0.9155** | **0.0011** |

Table 2 summarizes the ResNet18 image-classification results on CIFAR-10 and CIFAR-100. FLaG achieves the highest mean top-1 accuracy with 96.01% on CIFAR-10 and the second highest mean top 1 accuracy with 77.20% on CIFAR-100. It also obtains the highest average accuracy across the four reported metrics, at 91.55%, together with the lowest average standard deviation of 0.11 percentage points. These aggregate results indicate that FLaG provides the strongest overall accuracy profile and the lowest average across-seed dispersion among the evaluated pooling methods.

On CIFAR-10, FLaG reaches 96.01% top-1 accuracy and exceeds latent-attention, the strongest baseline on this metric, by 0.05 percentage points. The numerical separation among pooling methods is small, with mean top-1 accuracies ranging from 95.77% to 96.01%. Top-5 accuracy shows even less variation, with all methods between 99.86% and 99.88%. MLTP pooling achieve the highest mean top-5 accuracy of 99.88%, while FLaG reaches 99.86%. The CIFAR-10 results therefore show a modest top-1 advantage for FLaG, whereas top-5 accuracy remains nearly indistinguishable across pooling strategies at the reported precision.

The numerical separation is larger on CIFAR-100. FLaG achieves 77.20% top-1 accuracy, ranking second behind MLTP at 77.59% and exceeding mean pooling at 76.78%. FLaG also exceeds attention pooling by 4.15 percentage points and max pooling by 0.59 percentage points. Its top-1 standard deviation is 0.14 percentage points, which is the lowest among all evaluated methods on CIFAR-100. The corresponding values are 0.23 for attention pooling, 0.31 for mean pooling, 0.36 for max pooling, and 0.32 for MLTP. FLaG therefore combines the second-highest top-1 accuracy with the smallest across-seed dispersion in this setting.

The top-5 results follow a different pattern. On CIFAR-100, max pooling achieves the highest mean top-5 accuracy at 93.42%, followed by mean pooling at 93.25%, while FLaG reaches 93.13%. Thus, the dataset-specific advantage of FLaG is concentrated in top-1 accuracy rather than top-5 accuracy. However, the aggregate columns provide a broader view across both datasets and both accuracy measures. FLaG achieves an Avg. Acc. of 91.55%, compared with 91.46% for mean pooling and 91.45% for MLTP, which are the next strongest baselines. It also achieves the lowest Avg. Std. at 0.11 percentage points, compared with 0.12 for attention pooling and 0.15 for mean pooling. These results show that FLaG provides the highest overall mean accuracy while maintaining the lowest average variability across the four image-classification evaluations.

Overall, the image experiments show that FLaG does not lead every individual metric, particularly top-5 accuracy. Its advantage is instead reflected in strong top-1 performance on both datasets, a larger gain over conventional pooling methods on CIFAR-100, the highest aggregate accuracy, and the lowest average across-seed dispersion. This pattern supports the effectiveness of frequency-domain aggregation for visual representations under the evaluated ResNet18 setting.

### 4.2.3 Language Benchmarks

**Table 3. Language-task performance of pooling methods with RoBERTa-base on IMDB, SprintDuplicateQuestions, and STSBenchmark.** Results are reported as mean ± standard deviation over 10 random seeds. Bold values indicate the best mean in each metric column. Higher values are better for all reported metrics. IMDB and STSBenchmark use end-to-end adaptation, whereas SprintDuplicateQuestions uses a frozen RoBERTa backbone with adaptation of the pooling module and prediction head.

| Pooling | IMDB | SprintDuplicateQuestions | STSBenchmark |
|---|---|---|---|

| | Acc | F1 | Acc | Average Precision | F1 | Spearman | Pearson |
|---|---|---|---|---|---|---|---|
| mean | 0.9494±0.004 | 0.9501±0.004 | 0.9901±0.000 | 0.4288±0.000 | 0.0000±0.000 | **0.8515±0.003** | **0.8561±0.002** |
| max | 0.9471±0.007 | 0.9477±0.006 | 0.9901±0.000 | 0.4503±0.000 | 0.0000±0.000 | 0.8008±0.005 | 0.8084±0.004 |
| attn | 0.9104±0.004 | 0.9048±0.005 | 0.9701±0.006 | 0.3569±0.068 | 0.3001±0.070 | 0.6999±0.004 | 0.6836±0.005 |
| last | 0.9499±0.003 | 0.9502±0.003 | 0.9901±0.000 | 0.3020±0.000 | 0.0000±0.000 | 0.8248±0.005 | 0.8297±0.005 |
| MLTP | 0.9434±0.003 | 0.9455±0.003 | 0.9901±0.004 | 0.0119±0.000 | 0.000±0.000 | 0.3705±0.007 | 0.4180±0.005 |
| latent_attn | 0.9464±0.008 | 0.9471±0.007 | 0.9900±0.000 | 0.0409±0.011 | 0.0046±0.010 | 0.7594±0.008 | 0.7198±0.012 |
| first | 0.9483±0.005 | 0.9492±0.006 | 0.9901±0.000 | 0.2468±0.000 | 0.0000±0.000 | 0.8374±0.003 | 0.8401±0.003 |
| FLaG | **0.9516±0.002** | **0.9522±0.002** | **0.9930±0.001** | **0.6470±0.032** | **0.6169±0.026** | 0.8368±0.004 | 0.8315±0.005 |

Table 3 compares pooling strategies with RoBERTa-base on IMDB, SprintDuplicateQuestions, and STSBenchmark. The three tasks cover sentiment classification, duplicate-question detection, and semantic textual similarity. FLaG achieves the best mean result on five of the seven reported metrics, with its clearest advantage on SprintDuplicateQuestions.

On IMDB, the pooling methods perform within a narrow numerical range after end-to-end task adaptation. FLaG nevertheless achieves the highest accuracy and F1 score, both at 0.952±0.002. The strongest competing accuracy is 0.950 from last-token pooling, while several baselines reach an F1 score of 0.950. FLaG therefore improves both metrics by approximately 0.002 over the strongest baseline values. It also has the smallest standard deviation in both IMDB metrics. First-token pooling remains competitive at 0.948 accuracy and 0.949 F1, but it does not match FLaG.

The separation is substantially larger on SprintDuplicateQuestions. In this experiment, the RoBERTa backbone is frozen, while the pooling module and lightweight prediction head are adapted. FLaG achieves the highest accuracy, Average Precision, and F1 score, with values of 0.993±0.001, 0.647±0.032, and 0.617±0.026, respectively. Relative to the strongest non-FLaG result in each column, FLaG improves accuracy by 0.003, Average Precision by 0.197, and F1 by 0.317.

The SprintDuplicateQuestions results also reveal a substantial divergence between accuracy and the metrics that reflect positive-class recovery. Mean, max, last-token, and first-token pooling all achieve an accuracy of 0.990 but an F1 score of 0.000. This pattern is consistent with majority-class prediction under the adopted decision rule. Latent-attention shows a similar pattern, with 0.990 accuracy but an F1 score of only 0.005. FLaG instead maintains high accuracy together with substantially higher Average Precision and F1. These results indicate that its advantage in this task is not explained by accuracy alone. Prior work has likewise shown that token-level learning objectives can materially affect sentence representations [48], while token-weighted pooling can improve representations obtained from pretrained encoders without additional fine-tuning [49].

STSBenchmark produces a different ordering. Mean pooling achieves the highest Spearman and Pearson correlations at 0.852±0.003 and 0.856±0.002, respectively. FLaG reaches 0.837±0.004 for Spearman and 0.832±0.005 for Pearson. First-token pooling matches FLaG in Spearman at 0.837 and gives a higher Pearson correlation of 0.840. These results show that the benefit of FLaG does not extend uniformly to semantic textual similarity. In this setting, simple mean aggregation remains the strongest strategy. This observation is consistent with broad MTEB evaluations, which report substantial task dependence among text embedding approaches [30], and with prior work that shows the importance of token weighting for STS representations [49].

Overall, FLaG ranks first on five of the seven language metrics. It provides modest improvements on IMDB and substantially larger gains on SprintDuplicateQuestions, but mean pooling remains superior on STSBenchmark. The language results therefore support a task-dependent conclusion. Spectral aggregation is particularly effective for the evaluated classification and duplicate-detection settings, while uniform averaging remains preferable for semantic similarity under the present RoBERTa adaptation protocol.

### 4.2.4 Component Ablation Study

**Table 4. Component ablation of FLaG on AMP activity prediction with ESM2-8M.** Results are reported as mean ± standard deviation over 10 aligned random seeds. FFT+gate denotes the reduced variant that retains spectral transformation and gating without latent-attention. Bold values indicate the best mean in each row, and underlined values indicate the second-best mean. The Average rows report macro-averages across *E. coli* and *S. aureus*. Additional architectural variants and latent-dimension analyses are reported in Supplementary Material S2.

| Task | mean | max | latent_attn | FFT+gate | FLaG |
|---|---|---|---|---|---|
| *E. coli* RMSE ↓ | 0.5781±0.01 | 0.5765±0.03 | 0.5766±0.02 | **0.5530±0.01** | 0.5622±0.01 |
| *E. coli* Spearman ↑ | 0.5416±0.02 | 0.5404±0.02 | 0.5408±0.02 | **0.5611±0.01** | 0.5540±0.01 |
| *E. coli* Recall@50 ↑ | 17.7000±2.26 | 17.0000±2.49 | 16.7000±3.71 | **18.8000±2.30** | 18.6000±2.50 |
| *S. aureus* RMSE ↓ | 0.5773±0.02 | 0.5518±0.02 | 0.5699±0.01 | 0.5555±0.01 | **0.5449±0.00** |
| *S. aureus* Spearman ↑ | 0.5697±0.03 | 0.5909±0.02 | 0.5696±0.02 | 0.5846±0.02 | **0.6028±0.02** |
| *S. aureus* Recall@50 ↑ | 16.2000±2.82 | **18.4000±2.12** | 15.5000±2.17 | 18.1000±2.77 | 18.0000±1.76 |
| Avg. RMSE Mean↓ | 0.5777 | 0.5641 | 0.5733 | 0.5543 | **0.5535** |
| Avg. Spearman Mean↑ | 0.5556 | 0.5656 | 0.5552 | 0.5728 | **0.5784** |
| Avg. Recall@50 Mean↑ | 16.9500 | 17.7000 | 16.1000 | **18.4500** | 18.3000 |

Table 4 evaluates the contribution of the principal FLaG components on the ESM2-8M AMP tasks. In addition to mean, max, and latent-attention pooling, we include FFT+gate as the strongest reduced FLaG variant identified in the broader five-seed supplementary ablation. This variant retains spectral transformation and gating but removes latent-attention summarization. The main comparison uses 10 aligned random seeds.

FLaG achieves the best macro-averaged RMSE and Spearman correlation across the two species, with values of 0.5535 and 0.5784, respectively. Relative to max pooling, the strongest conventional baseline in the aggregate comparison, FLaG reduces average RMSE from 0.5641 to 0.5535 and increases average Spearman correlation from 0.5656 to 0.5784. FLaG also improves both aggregate metrics over latent-attention alone. These comparisons show that neither latent-attention nor conventional token pooling reproduces the performance of the complete spectral architecture.

The contribution of latent-attention differs between the two species. On *E. coli*, the reduced FFT+gate variant performs best on all three metrics. It achieves an RMSE of 0.5530, a Spearman correlation of 0.5611, and Recall@50 of 18.8. FLaG reaches 0.5622, 0.5540, and 18.6, respectively. On *S. aureus*, the full FLaG architecture provides a clearer advantage in regression and ranking performance. It achieves the lowest RMSE of 0.5449 and the highest Spearman correlation of 0.6028. Compared with FFT+gate, FLaG reduces RMSE by 0.0106 and increases Spearman correlation by 0.0182. These results indicate that latent spectral summarization provides a task-dependent contribution rather than a uniform improvement across both AMP datasets.

Recall@50 follows a different pattern. FFT+gate achieves the highest macro-average Recall@50 at 18.45, followed by FLaG at 18.30. At the species level, FFT+gate leads on *E. coli*, whereas max pooling gives the highest *S. aureus* Recall@50 at 18.4. The configuration that minimizes regression error and improves global rank correlation therefore does not necessarily maximize recovery among the top 50 candidates.

Taken together, the ablation results support complementary roles for the FLaG components. Spectral transformation combined with gating already provides a strong reduced architecture, particularly for *E. coli*. The addition of latent-attention improves the macro-averaged RMSE and Spearman correlation because of its stronger performance on *S. aureus*, but it does not improve every species-metric combination. Unlike the reduced FFT+gate variant, full FLaG contains a latent-query cross-attention module that exposes query-conditioned spectral attention weights for examining species-specific frequency utilization. Therefore, when macro-averaged regression and ranking performance are considered together with the capacity for mechanism analysis, FLaG provides a more complete solution despite the advantages of FFT+gate in several individual settings.  This result also motivates the species-specific mechanism analyses in the following sections, which examine how FLaG uses spectral information differently for *E. coli* and *S. aureus*.

The broader five-seed component study is reported in Supplementary Material S2. It includes alternative final pooling operators, removal of residual gating, additional combinations of the FFT, latent-attention, and gating components, and the latent-dimension analysis. In that analysis, (L=8) provides the best overall balance between RMSE and Spearman correlation, whereas smaller latent dimensions can yield higher Recall@50.

## 4.3 AMP-Side Mechanistic Analysis

AMP activity prediction provided the original biological motivation for FLaG and showed strong aggregate improvements in the preceding experiments. It also provides residue-level and physicochemical information that allows the pooling mechanism to be examined at multiple levels. However, learned attention patterns should not be interpreted directly as biological explanations. Cross-attention queries are optimized to retrieve information useful for downstream prediction, and attention weights do not necessarily identify features with causal importance [35]. We therefore use complementary perturbation and representation analyses to characterize the behavior of FLaG.

We conduct five AMP-side analyses with ESM2-8M on both *E. coli* and *S. aureus*: single-position hidden-state knockout, sequence-frequency band knockout, gate spectral-effect analysis, latent-query cross-attention readout, and a two-part physicochemical-property analysis. The first physicochemical part examines property encoding and property-stratified spectral sensitivity, whereas the second tests species-specific associations between these properties and antimicrobial activity. Together, the five analyses examine positional sensitivity, frequency dependence, spectral modulation, latent spectral allocation, and the physicochemical content and activity relevance of spectral representations.

### 4.3.1 Single-Position Hidden-State Knockout Reveals Positional Response Heterogeneity

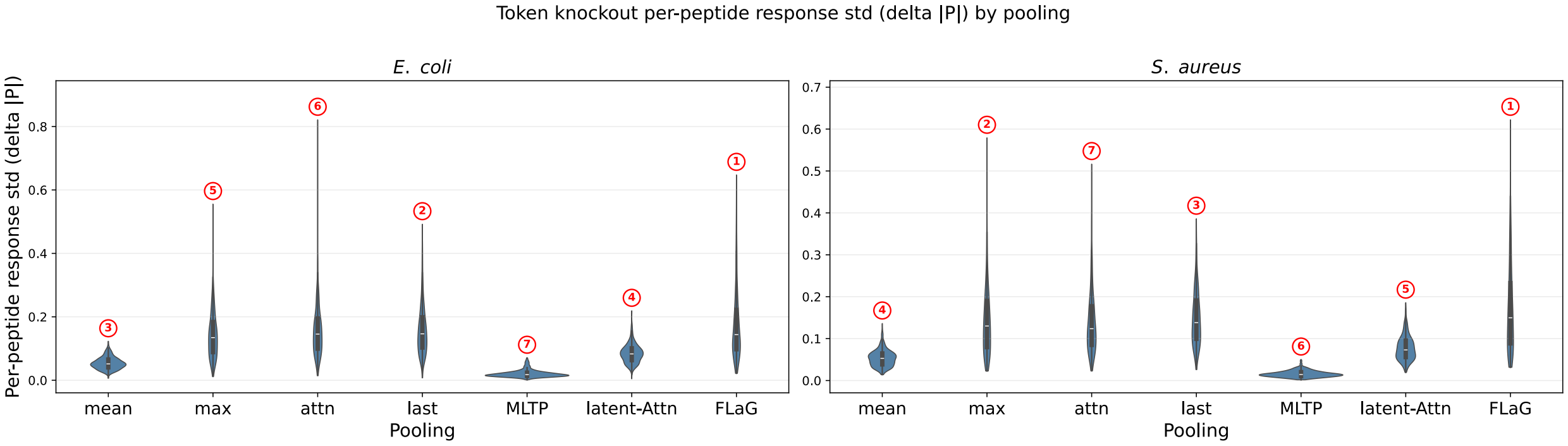

**Figure 2. Within-peptide positional response heterogeneity under single-position hidden-state knockout across pooling methods.** The analysis uses ESM2-8M on the full *E. coli* (left) and *S. aureus* (right) test sets. For each peptide, the final-layer hidden vector at each valid residue position is individually set to zero, and the absolute squared prediction error change in the predicted activity score on the $-\log_{10}(\mathrm{MIC}[M])$ regression scale, $|\Delta \mathrm{P}|$, is averaged over 10 aligned training seeds. The standard deviation of these responses across residue positions defines the per-peptide positional-response heterogeneity. Each point in a violin represents one peptide. The evaluated methods are mean, max, attn, last, MLTP, latent_attn, and FLaG. Red circled numbers indicate the test-set Spearman rank of each pooling method for the corresponding species, as reported in Table 1. Larger response standard deviations indicate stronger variation in prediction sensitivity across residue positions.

We first examine whether prediction sensitivity varies across residue positions within individual peptides. For each peptide, we set the final-layer ESM2 hidden vector at one valid residue position to zero while leaving all other hidden vectors unchanged. We then recompute the pooled representation and activity prediction. The response at each position is defined as the absolute change in squared prediction error after perturbing the predicted activity score (similar to Eq. 15; see Supplementary Material S3.1 for details). For each position, this response is averaged over 10 trained models from the aligned random seeds. We then calculate the standard deviation of the averaged responses across valid residue positions within each peptide. Each peptide therefore contributes one positional-response standard deviation. A larger value indicates greater variation in model sensitivity across residue positions, whereas a smaller value indicates a more uniform response. Figure 2 summarizes these distributions.

Figure 2 reveals clear differences in positional-response heterogeneity among pooling methods. FLaG exhibits a broad distribution toward larger per-peptide response standard deviations for both *E. coli* and *S. aureus*. Its upper range is particularly pronounced for *S. aureus*, whereas attention reaches the largest values for *E. coli*, FLaG reaches the second one. As indicated by the circled ranks, it ranks first in Spearman correlation among the ESM2-8M pooling methods for both species. Thus, the improved predictive performance of FLaG coincides with less uniform sensitivity across residue positions within individual peptides.

Max, attention and last-token pooling also produce relatively broad response distributions, but their predictive ranks remain below FLaG. In contrast, mean pooling and MLTP exhibit narrower distributions, which indicate more uniform positional responses for most peptides. Latent-attention show intermediate patterns. The qualitative relationship is similar across the two bacterial species, although the magnitude and spread of the responses vary between methods and species.

These results indicate that FLaG differentiates residue positions more strongly at the level of prediction sensitivity than mean pooling and MLTP, while showing sensitivity heterogeneity comparable to other position-selective pooling strategies. They do not establish that greater heterogeneity itself causes higher predictive performance. The knockout also perturbs contextual hidden representations rather than the peptide sequence directly. The observed responses should therefore be interpreted as model-level positional sensitivity, not as direct evidence of biological residue essentiality. The following frequency-domain analyses examine whether this differentiated sensitivity is associated with specific spectral components of the encoded peptide representation.

### 4.3.2 Sequence-Frequency Band Knockout Reveals a Layer-Wise Low-to-High Frequency Shift

**Table 5. Examples of the eight-band DCT partition for sequence lengths $T = 9$ and $T = 25$.** Frequency indices are divided into eight contiguous, non-overlapping half-open intervals using the geometric allocation described above. $|\boldsymbol{\mathcal{B}}_b|$ denotes the number of DCT coefficients assigned to band $b$. The partition requires $T \geq K = 8$.

| seq length | 9 | | 25 | |
|---|---|---|---|---|
| **b** | $\boldsymbol{\mathcal{B}}_b$ | $\lvert\boldsymbol{\mathcal{B}}_b\rvert$ | $\boldsymbol{\mathcal{B}}_b$ | $\lvert\boldsymbol{\mathcal{B}}_b\rvert$ |
| **0** | [0,1) | 1 | [0,1) | 1 |
| **1** | [1,2) | 1 | [1,2) | 1 |
| **2** | [2,3) | 1 | [2,3) | 1 |
| **3** | [3,4) | 1 | [3,4) | 1 |
| **4** | [4,5) | 1 | [4,5) | 1 |
| **5** | [5,6) | 1 | [5,7) | 2 |
| **6** | [6,7) | 1 | [7,11) | 4 |
| **7** | [7,9) | 2 | [11,25) | 14 |

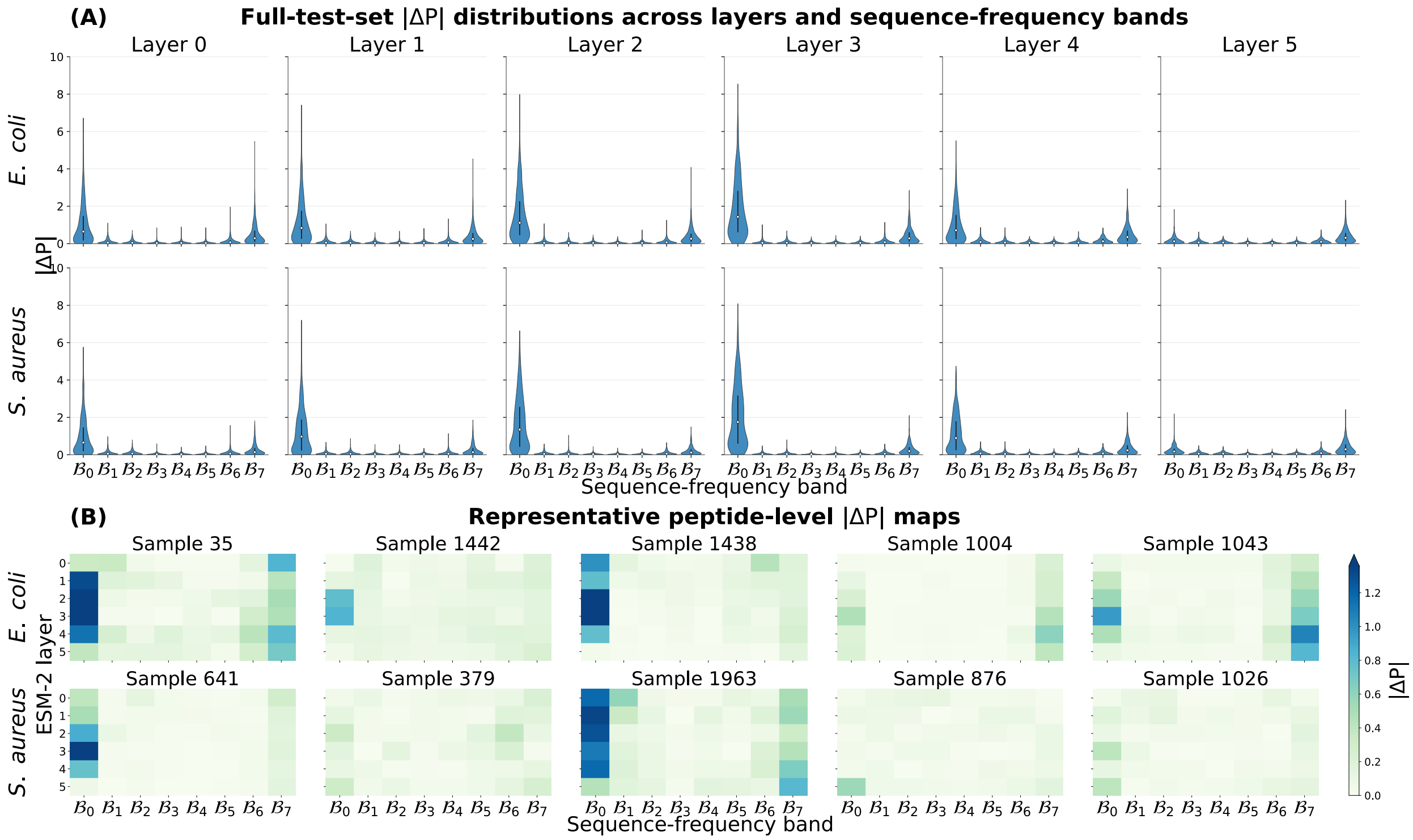

**Figure 3. Layer-wise sensitivity to sequence-frequency band knockout in ESM2-8M.** (A) Full-test-set distributions of the per-peptide absolute change in squared prediction error, $|\Delta P_{i,l,b}|$, after removal of one DCT frequency band. Results are shown across the six ESM2-8M transformer layers and eight sequence-frequency bands for *E. coli* (top) and *S. aureus* (bottom). Responses are averaged over 10 aligned training seeds before aggregation across peptides. Each violin represents one layer-band combination. Larger values indicate greater sensitivity of prediction error to removal of the corresponding band. (B) Representative peptide-level heatmaps of ($|\Delta P_{i,l,b}|$). Rows denote encoder layers, and columns denote frequency bands from $\mathcal{B}_0$ to $\mathcal{B}_7$. Darker values indicate greater sensitivity to band removal. The color scale is clipped at the 99th percentile of all displayed |ΔP| values; the triangular tip of the colorbar marks entries exceeding this threshold, which are shown with the maximum color.

The previous section showed that FLaG exhibits nonuniform sensitivity across residue positions. We next examine whether this sensitivity can be further resolved according to sequence-frequency scale and encoder depth. We perform a per-peptide, band-wise knockout analysis on the hidden representations of ESM2-8M for both *E. coli* and *S. aureus*.

For this analysis, we adapt the DCT-based spectral filtering framework [24] rather than directly perturbing the rFFT coefficients inside FLaG. At a selected encoder layer, the hidden representation is transformed along the sequence axis with the discrete cosine transform (DCT). One frequency band is then removed with a binary notch mask, and the perturbed representation is reconstructed with the inverse DCT. FLaG itself continues to use rFFT during model inference. The DCT is used only as a real-valued analysis basis for controlled frequency-band perturbation.

Let $\mathbf{H}_{\mathrm{i}}^{l} \in \mathbb{R}^{T_i \times D}$ denote the hidden representation of peptide $i$ at encoder layer $l$, where $T_i$ is the sequence length used in the spectral analysis and $D$ is the hidden dimension. We apply the DCT independently along the sequence axis and obtain $\mathbf{Z}_i^{(l)} = \mathrm{DCT}(\mathbf{H}_i^{l}) \in \mathbb{R}^{T_i \times D}$.

The $T_i$ DCT coefficients are partitioned into $K = 8$ contiguous and non-overlapping frequency bands, denoted by $\boldsymbol{\mathcal{B}}_0, \dots, \boldsymbol{\mathcal{B}}_7$. We use a geometric allocation with base 4. Each band receives at least one coefficient, and the remaining coefficients are allocated with weights proportional to $4^b$. Consequently, $\boldsymbol{\mathcal{B}}_0$ contains the lowest-frequency components, which vary most slowly along the sequence axis, while $\boldsymbol{\mathcal{B}}_7$ contains the highest-frequency components, which vary most rapidly. Table 5 gives two examples of the resulting partition.

Let $m_b \in \{0,1\}^{T_i}$ denote the binary mask for band $b$, with entries equal to zero for indices in $B_b$ and one elsewhere. The mask is broadcast across the hidden dimension. The masked coefficients and their initial reconstruction are

$$\mathbf{Z}_{i,l,b}^{ko} = \mathbf{Z}_i^{(l)} \odot m_b, \qquad \widetilde{\mathbf{H}}_{i,l,b} = i\mathrm{DCT}(\mathbf{Z}_{i,l,b}^{ko}). \tag{14}$$

Band removal can alter the magnitude of individual token representations.

Let $y_i$ denote the observed activity, $\hat{y}_i$ the baseline prediction, and $\hat{y}_{i,l,b}^{\mathrm{ko}}$ the prediction after band $b$ is removed at layer $l$. Because the response is defined for an individual peptide, we quantify the perturbation as the absolute change in squared prediction error:

$$|\Delta \mathrm{P}_{i,\ell,b}| = |(\hat{y}_{i,\ell,b}^{\mathrm{ko}} - \mathrm{y})^2 - (\hat{y}_i - \mathrm{y})^2|. \tag{15}$$

The response is computed for each of the 10 aligned training seeds and then averaged for each peptide, layer, and band. Figure 3 summarizes the resulting $|\Delta \mathrm{P}_{i,\ell,b}|$ values across both AMP test sets.

Figure 3A reveals a consistent depth-dependent shift in frequency sensitivity across the two AMP datasets. In shallow ESM2-8M layers, predictions are most sensitive to removal of the lowest-frequency band $\boldsymbol{\mathcal{B}}_{\mathbf{0}}$. This result indicates stronger predictive dependence on slowly varying sequence components at early encoder depths. In the final layer, the sensitivity to $\boldsymbol{\mathcal{B}}_{\mathbf{0}}$ decreases substantially, while the relative sensitivity to the highest-frequency band $\boldsymbol{\mathcal{B}}_{\mathbf{7}}$ becomes more pronounced. Deeper representations therefore show an increased relative dependence on rapidly varying sequence components rather than a complete replacement of low-frequency sensitivity. Intermediate-frequency bands exhibit smaller changes across depth and do not show the same systematic transition. This low-to-high frequency pattern qualitatively resembles prior observations that shallow network layers preferentially learn lower-frequency target components, whereas deeper layers capture higher-frequency components [50].

Figure 3B shows that the aggregate depth-dependent pattern does not imply an identical frequency profile for every peptide. Representative samples exhibit distinct layer-band response patterns, especially at the lowest and highest frequency bands. The middle-frequency bands are generally less affected by depth. Additional peptide-level examples are provided in Supplementary Material S3.2.

Together, these results identify a layer-dependent redistribution of predictive frequency sensitivity in the ESM2-8M representations. Slowly varying sequence components dominate the absolute response at shallow depths, while rapidly varying components make a larger relative contribution in deeper layers. The analysis does not imply that the corresponding DCT components are biologically causal or that they are identical to the internal rFFT coefficients used by FLaG. Instead, it provides an external spectral probe of how predictive information is distributed across encoder depth. This result motivates the following analyses of FLaG's learned spectral gate and latent-query allocation.

### 4.3.3 Latent-Query Readout Reveals Sample- and Species-Specific Spectral Allocation

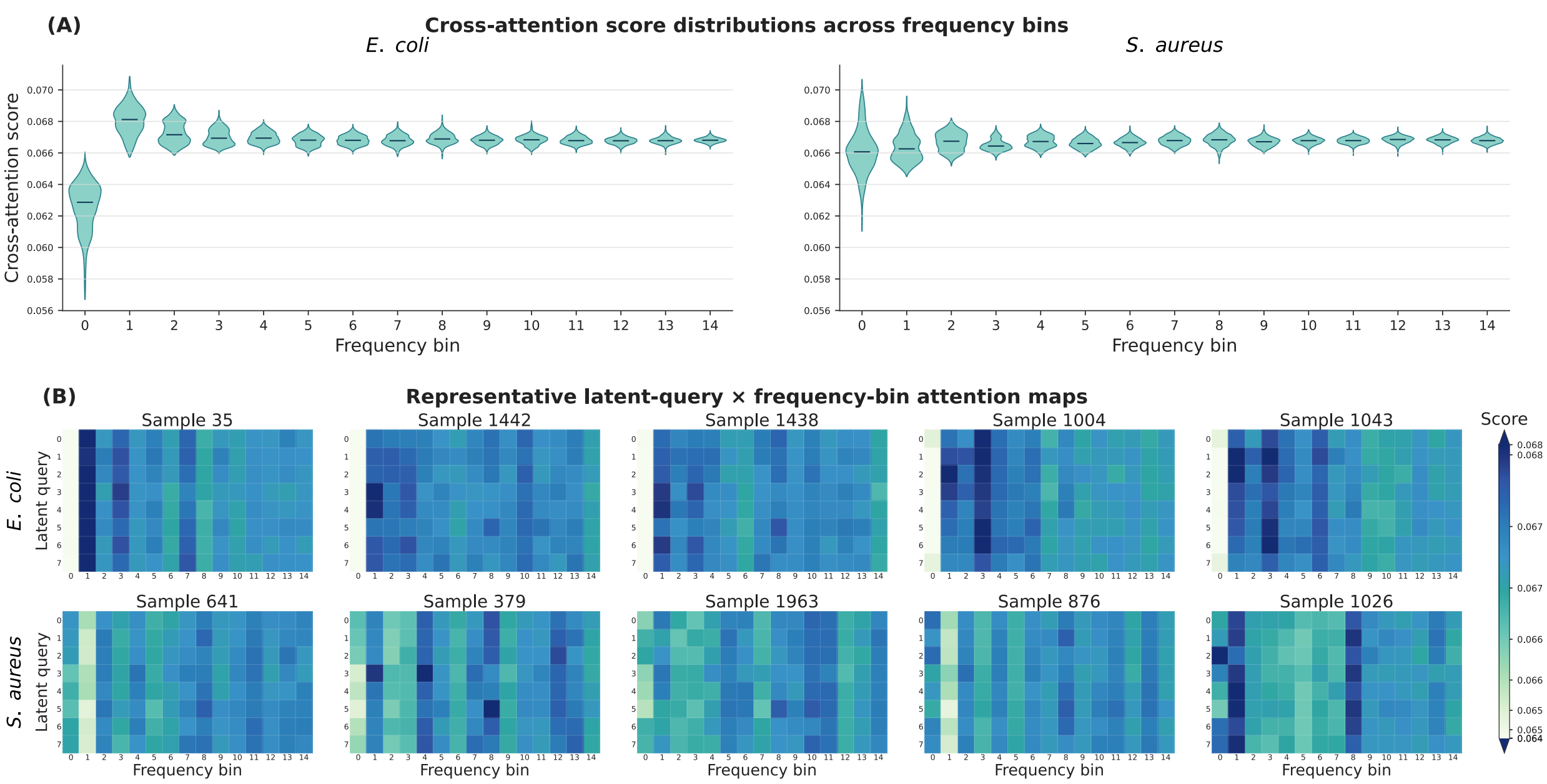

Figure 4. Latent-query cross-attention across rFFT frequency bins in FLaG. (A) Post-softmax Cross-attention distributions across 15 nonredundant rFFT bins for the full E. coli and S. aureus test sets. Weights are averaged over four heads and 10 aligned training seeds. Each violin pools all eight latent queries and test peptides, with inner boxes showing the median and interquartile range. Bin 0 is the DC component and bin 1 is the first non-DC component. E. coli shows lower attention at bin 0 and higher attention at bin 1, while this contrast is weaker for S. aureus. (B) Representative peptide-level attention maps for five peptides per task. The color scale is clipped at the 99th percentile, with larger values shown using the maximum color. Additional examples are provided in Supplementary Material S3.4.

The preceding analysis showed that FLaG gating amplifies spectral channels while retaining a low-frequency-dominated energy profile. Because the channel gate is shared across frequency bins, we examined whether the latent cross-attention stage provides frequency-specific allocation using the weights from Eq. 3.

Figure 4A shows a distinct low-frequency pattern for *E. coli*. The DC component at bin 0 receives relatively low attention, whereas the first non-DC component at bin 1 receives the largest attention among the first several bins. Attention becomes comparatively stable from bin 2 onward. Latent-attention is therefore redistributed from the global DC component toward the lowest nonzero sequence frequency. For *S.*

*aureus*, the contrast between bins 0 and 1 is weaker, their distributions overlap more with the remaining bins, and DC attention varies more across peptides.

Figure 4B confirms this task difference at the peptide level. Bin 0 commonly receives less attention than bin 1 for *E. coli*, although the magnitude varies among samples. The *S. aureus* examples show less consistent relationships. Some peptides retain strong DC attention, and local maxima can occur at intermediate frequencies. Supplementary Material S3.4 shows similar peptide-to-peptide variation.

The eight latent queries show modest within-peptide differences but generally share the broader sample-level spectral profile. This pattern is consistent with sample-conditioned latent readout rather than fixed assignment of queries to frequency regions. The ablation in Supplementary Material S2 shows that reducing the latent set to one query lowers predictive performance. This supports multiple latent slots but does not establish that the visible differences among queries cause the gain.

Together with Figure S2, these results show that latent-attention does not simply reproduce spectral energy. Although DC has the largest frequency-token energy, *E. coli* queries allocate less attention to DC and more to the first non-DC bin. This allocation is less consistent across *S. aureus* peptides. These patterns describe model-level allocation rather than biological importance. Cross-attention retrieves information useful for prediction [37], but its weights do not establish causal feature importance [35]. The next analysis therefore examines the physicochemical properties encoded by these spectral components and their species-specific associations with antimicrobial activity.

### 4.3.4 Spectral Gating and Physicochemical Properties of Low-Frequency Representations

We next examined how FLaG gating affects the spectral representation and what physicochemical information is associated with its low-frequency components. Following the layer-wise analysis above, we first compared frequency-token energy before and after gating across eight spectral bands in the E. coli and S. aureus test sets. Because the residual gate scales spectral channels by factors greater than one, increased energy after gating is expected by construction. We therefore focused on whether gating altered the relative spectral profile. In both tasks, the lowest-frequency band remained dominant after gating, with no clear redistribution toward intermediate- or high-frequency components. This pattern is consistent with broad channel-wise recalibration rather than selective enhancement of particular frequency bands. Detailed definitions, procedures, and results are provided in Supplementary Section S3.3.

Given the persistent low-frequency dominance, we next examined which physicochemical properties were decodable from these representations. Ridge probes were used to decode seven sequence-derived properties from individual final-layer DCT components, with amino-acid composition as a composition-only reference. Across both organisms, all properties were most strongly decodable from the normalized DCT DC component. This included the composition-derived helix-group fraction. Hydrophobic-moment decoding also suggested that the contextualized DC representation contains information beyond residue frequencies alone. Property-stratified knockout analysis further showed that prediction sensitivity was mainly concentrated in the lowest-frequency band, although the patterns differed across properties and organisms. Full descriptor definitions, probe procedures, bootstrap analyses, and stratified results are provided in Supplementary Section S3.5 and the corresponding results in Supplementary Section S3.6.

We further tested whether these physicochemical properties had different associations with antimicrobial activity across species. A pooled regression analysis examined four properties while accounting for peptide length. Net charge was more strongly associated with E. coli activity, whereas mean hydrophobicity was more strongly associated with S. aureus activity. No significant species-specific differences were observed for hydrophobic moment or helix-group fraction. These results suggest that similar global physicochemical properties are decodable from the low-frequency representations in both tasks, while their associations with antimicrobial activity differ between the two bacterial species. Further details and results of the species-specific analysis are provided in Supplementary Sections S3.5.4 and S3.7.

# 5 Conclusions

We introduced FLaG, a plug-in token aggregation module that performs spectral re-expression, latent-query summarization, sample-conditioned channel gating, inverse reconstruction, and token pooling. Evaluated across AMP activity prediction, image classification, and language tasks, FLaG achieved the best aggregate Spearman correlation, RMSE, and Recall@50 across four AMP backbone-species settings, with its clearest advantage under ESM2-8M. It also obtained the highest top-1 accuracy on CIFAR-10 and performed best on IMDB and SprintDuplicateQuestions, but remained below mean pooling on STSBenchmark. Ablations showed that the complete model achieved the best average AMP RMSE and Spearman correlation, whereas the reduced FFT+gate variant achieved the highest average Recall@50, indicating that latent spectral summarization provides task-dependent benefits.

AMP analyses showed heterogeneous positional responses, predominant low-frequency sensitivity across most ESM2-8M layers, with increased relative high-frequency sensitivity in the final layer, and broad sample-conditioned channel modulation by the residual gate. Latent cross-attention produced more frequency-specific and species-dependent allocation. The normalized DCT DC representation strongly decoded global physicochemical properties in both organisms, while hydrophobic-moment results indicated sequence-organization information beyond amino-acid composition. Species-interaction analysis further supported shared low-frequency content with differential relevance to antimicrobial activity.

Overall, spectral re-expression provides a transferable aggregation coordinate system, but FLaG is not universally optimal. Its benefits reflect interactions among global spectral content, non-DC components, encoder depth, positional responses, and learned allocation rather than simple low-frequency preservation. Current limitations include the absence of spectral localization, mechanistic evaluation restricted to AMP tasks with ESM2-8M, and unresolved interpretations of non-DC components. Local time-frequency or adaptive spectral methods [51, 52], controlled sequence interventions, and structure-aware validation may clarify these components and test whether the observed mechanisms generalize across modalities.

## Acknowledgements

This study was supported by the Fundamental Research Funds for the Central Universities (JLU).

## References

[1] H. Lee, S. Lee, I. Lee, and H. Nam, "AMP-BERT: Prediction of antimicrobial peptide function based on a BERT model," *Protein Science,* vol. 32, no. 1, p. e4529, 2023.

[2] N. NaderiAlizadeh and R. Singh, "Aggregating residue-level protein language model embeddings with optimal transport," *Bioinformatics Advances,* vol. 5, no. 1, p. vbaf060, 2025.

[3] A. Rives *et al.*, "Biological structure and function emerge from scaling unsupervised learning to 250 million protein sequences," *Proceedings of the national academy of sciences,* vol. 118, no. 15, p. e2016239118, 2021.

[4] A. Dosovitskiy *et al.*, "An image is worth 16x16 words: Transformers for image recognition at scale," in *International Conference on Learning Representations*, 2021.

[5] J. Devlin, M.-W. Chang, K. Lee, and K. Toutanova, "Bert: Pre-training of deep bidirectional transformers for language understanding," in *Proceedings of the 2019 conference of the North American chapter of the association for computational linguistics: human language technologies, volume 1 (long and short papers)*, 2019, pp. 4171-4186.

[6] M. Zaheer, S. Kottur, S. Ravanbakhsh, B. Poczos, R. R. Salakhutdinov, and A. J. Smola, "Deep sets," *Advances in neural information processing systems,* vol. 30, 2017.

[7] J. Lee, Y. Lee, J. Kim, A. Kosiorek, S. Choi, and Y. W. Teh, "Set transformer: A framework for attention-based permutation-invariant neural networks," in *International conference on machine learning*, 2019: PMLR, pp. 3744-3753.

[8] N. Reimers and I. Gurevych, "Sentence-bert: Sentence embeddings using siamese bert-networks," in *Proceedings of the 2019 conference on empirical methods in natural language processing and the 9th international joint conference on natural language processing (EMNLP-IJCNLP)*, 2019, pp. 3982-3992.

[9] Z. Lin *et al.*, "A structured self-attentive sentence embedding," in *Proceedings of the 5th International Conference on Learning Representations (ICLR)*, Toulon, France, 2017.

[10] L. Beyer, X. Zhai, and A. Kolesnikov, "Better plain vit baselines for imagenet-1k," *arXiv preprint arXiv:2205.01580,* 2022.

[11] D. Marin, J.-H. R. Chang, A. Ranjan, A. Prabhu, M. Rastegari, and O. Tuzel, "Token pooling in vision transformers for image classification," in *Proceedings of the IEEE/CVF winter conference on applications of computer vision*, 2023, pp. 12-21.

[12] Z. Lin *et al.*, "Evolutionary-scale prediction of atomic-level protein structure with a language model," *Science,* vol. 379, no. 6637, pp. 1123-1130, 2023.

[13] F. Wan, F. Wong, J. J. Collins, and C. de la Fuente-Nunez, "Machine learning for antimicrobial peptide identification and design," *Nature Reviews Bioengineering,* vol. 2, no. 5, pp. 392-407, 2024.

[14] K. He, X. Zhang, S. Ren, and J. Sun, "Deep residual learning for image recognition," in *Proceedings of the IEEE conference on computer vision and pattern recognition*, 2016, pp. 770-778.

[15] Y. Liu *et al.*, "Roberta: A robustly optimized bert pretraining approach," *arXiv preprint arXiv:1907.11692,* 2019.

[16] C. Lee *et al.*, "Nv-embed: Improved techniques for training llms as generalist embedding models," in *International Conference on Learning Representations*, 2025.

[17] Y. Tang and Y. Yang, "Pooling and attention: What are effective designs for llm-based embedding models?," *arXiv preprint arXiv:2409.02727,* 2024.

[18] A. Jaegle *et al.*, "Perceiver io: A general architecture for structured inputs & outputs," in *International Conference on Learning Representations*, 2022.

[19] A. Jaegle, F. Gimeno, A. Brock, O. Vinyals, A. Zisserman, and J. Carreira, "Perceiver: General perception with iterative attention," in *International conference on machine learning*, 2021: PMLR, pp. 4651-4664.

[20] J. Hyun, H. Seong, and E. Kim, "Universal pooling–a new pooling method for convolutional neural networks," *Expert Systems with Applications,* vol. 180, p. 115084, 2021.

[21] J. Lee-Thorp, J. Ainslie, I. Eckstein, and S. Ontanon, "Fnet: Mixing tokens with fourier transforms," in *Proceedings of the 2022 Conference of the north American chapter of the Association for Computational Linguistics: human language technologies*, 2022, pp. 4296-4313.

[22] N. Rahaman *et al.*, "On the spectral bias of neural networks," in *International conference on machine learning*, 2019: PMLR, pp. 5301-5310.

[23] T. Zhou, Z. Ma, Q. Wen, X. Wang, L. Sun, and R. Jin, "Fedformer: Frequency enhanced decomposed transformer for long-term series forecasting," in *International conference on machine learning*, 2022: PMLR, pp. 27268-27286.

[24] A. Tamkin, D. Jurafsky, and N. Goodman, "Language through a prism: A spectral approach for multiscale language representations," *Advances in Neural Information Processing Systems,* vol. 33, pp. 5492-5504, 2020.

[25] Y. Rao, W. Zhao, Z. Zhu, J. Zhou, and J. Lu, "GFNet: Global filter networks for visual recognition," *IEEE Transactions on Pattern Analysis and Machine Intelligence,* vol. 45, no. 9, pp. 10960-10973, 2023.

[26] N. NaderiAlizadeh and R. Singh, "EvoPool: Evolution-Guided Pooling of Protein Language Model Embeddings," *bioRxiv,* p. 2026.02. 02.703349, 2026.

[27] J. Hu, L. Shen, and G. Sun, "Squeeze-and-excitation networks," in *Proceedings of the IEEE conference on computer vision and pattern recognition*, 2018, pp. 7132-7141.

[28] K. Li *et al.*, "AMPCliff: Quantitative definition and benchmarking of activity cliffs in antimicrobial peptides," *Journal of Advanced Research,* vol. 80, pp. 287–300, 2026, doi: 10.1016/j.jare.2025.04.046.

[29] A. Krizhevsky and G. Hinton, "Learning multiple layers of features from tiny images," 2009.

[30] N. Muennighoff, N. Tazi, L. Magne, and N. Reimers, "Mteb: Massive text embedding benchmark," in *Proceedings of the 17th Conference of the European Chapter of the Association for Computational Linguistics*, 2023, pp. 2014-2037.

[31] A. Tartici, G. Nayar, and R. B. Altman, "Pool PaRTI: a PageRank-based pooling method for identifying critical residues and enhancing protein sequence representations," *Bioinformatics,* vol. 41, no. 6, p. btaf330, 2025.

[32] A. Radford, K. Narasimhan, T. Salimans, and I. Sutskever, "Improving language understanding by generative pre-training," 2018.

[33] M. Ilse, J. Tomczak, and M. Welling, "Attention-based deep multiple instance learning," in *International conference on machine learning*, 2018: PMLR, pp. 2127-2136.

[34] A. Sadafi, N. Navab, and C. Marr, "Active learning enhances classification of histopathology whole slide images with attention-based multiple instance learning," in *2023 IEEE 20th International Symposium on Biomedical Imaging (ISBI)*, 2023: IEEE, pp. 1-5.

[35] S. Jain and B. C. Wallace, "Attention is not explanation," in *Proceedings of the 2019 Conference of the North American Chapter of the Association for Computational Linguistics: Human Language Technologies, Volume 1 (Long and Short Papers)*, 2019, pp. 3543-3556.

[36] I. Tenney, D. Das, and E. Pavlick, "BERT rediscovers the classical NLP pipeline," in *Proceedings of the 57th annual meeting of the association for computational linguistics*, 2019, pp. 4593-4601.

[37] L. Feng, F. Tung, H. Hajimirsadeghi, Y. Bengio, and M. O. Ahmed, "Tree cross attention," in *International Conference on Learning Representations*, 2024.

[38] T. N. Kipf and M. Welling, "Semi-supervised classification with graph convolutional networks," in *Proceedings of the 5th International Conference on Learning Representations (ICLR)*, Toulon, France, 2017.

[39] Z. Xu, J. Wu, Y. S. Song, and R. Mahadevan, "Enzyme activity prediction of sequence variants on novel substrates using improved substrate encodings and convolutional pooling," in *Machine Learning in Computational Biology*, 2022: PMLR, pp. 78-87.

[40] A. Vaswani *et al.*, "Attention is all you need," *Advances in neural information processing systems,* vol. 30, 2017.

[41] J. L. Ba, J. R. Kiros, and G. E. Hinton, "Layer normalization," *arXiv preprint arXiv:1607.06450,* 2016.

[42] R. Xiong *et al.*, "On layer normalization in the transformer architecture," in *International conference on machine learning*, 2020: PMLR, pp. 10524-10533.

[43] J. Xu, X. Sun, Z. Zhang, G. Zhao, and J. Lin, "Understanding and improving layer normalization," *Advances in neural information processing systems,* vol. 32, 2019.

[44] L. Wang *et al.*, "Text embeddings by weakly-supervised contrastive pre-training," *arXiv preprint arXiv:2212.03533,* 2022.

[45] S. Sturua *et al.*, "jina-embeddings-v3: Multilingual embeddings with task lora," *arXiv preprint arXiv:2409.10173,* 2024.

[46] H. Su *et al.*, "One embedder, any task: Instruction-finetuned text embeddings," in *Findings of the Association for Computational Linguistics: ACL 2023*, 2023, pp. 1102-1121.

[47] A. Conneau, D. Kiela, H. Schwenk, L. Barrault, and A. Bordes, "Supervised learning of universal sentence representations from natural language inference data," in *Proceedings of the 2017 conference on empirical methods in natural language processing*, 2017, pp. 670-680.

[48] J. M. Janeiro, B. Piwowarski, P. Gallinari, and L. Barrault, "MEXMA: Token-level objectives improve sentence representations," in *Proceedings of the 63rd Annual Meeting of the Association for Computational Linguistics (Volume 1: Long Papers)*, 2025, pp. 23960-23995.

[49] Q. Chen *et al.*, "Ditto: A simple and efficient approach to improve sentence embeddings," in *Proceedings of the 2023 Conference on Empirical Methods in Natural Language Processing*, 2023, pp. 5868-5875.

[50] Y. Chen, A. Yuille, and Z. Zhou, "Which layer is learning faster? a systematic exploration of layer-wise convergence rate for deep neural networks," in *The Eleventh International Conference on Learning Representations*, 2023.

[51] I. Naiman, N. Berman, I. Pemper, I. Arbiv, G. Fadlon, and O. Azencot, "Utilizing image transforms and diffusion models for generative modeling of short and long time series," *Advances in Neural Information Processing Systems,* vol. 37, pp. 121699-121730, 2024.

[52] W. Ye, S. Deng, Q. Zou, and N. Gui, "Frequency adaptive normalization for non-stationary time series forecasting," *Advances in neural information processing systems,* vol. 37, pp. 31350-31379, 2024.